\documentclass{article}

\usepackage{graphicx}
\usepackage{amsmath}
\usepackage{amsfonts}

\usepackage{hyperref}

\usepackage[vlined,linesnumbered,boxed]{algorithm2e}

\usepackage[accepted]{icml2018}

\icmltitlerunning{Automatic formation of the structure of abstract machines in hierarchical reinforcement learning with state clustering}

\begin{document}

\twocolumn[
\icmltitle{Automatic formation of the structure of abstract machines in hierarchical reinforcement learning with state clustering}

\begin{icmlauthorlist}
\icmlauthor{Aleksandr I. Panov}{to}
\icmlauthor{Aleksey Skrynnik}{to}
\end{icmlauthorlist}

\icmlaffiliation{to}{Federal Research Center ``Computer Science and Control'' of the Russian Academy of Sciences, Moscow, Russia}

\icmlcorrespondingauthor{Aleksandr I. Panov}{pan@isa.ru}

\icmlkeywords{Reinforcement Learning, Hierarchical Reinforcement Learning, HAM, skill discovery, hierarchy formation}

\vskip 0.3in
]

\printAffiliationsAndNotice{}  

\begin{abstract}
We introduce a new approach to hierarchy formation and task decomposition in hierarchical reinforcement learning. Our method is based on the Hierarchy Of Abstract Machines (HAM) framework because HAM approach is able to design efficient controllers that will realize specific behaviors in real robots. The key to our algorithm is the introduction of the internal or ``mental'' environment in which the state represents the structure of the HAM hierarchy. The internal action in this environment leads to changes the hierarchy of HAMs. We propose the classical Q-learning procedure in the internal environment which allows the agent to obtain an optimal hierarchy. We extends the HAM framework by adding on-model approach to select the appropriate sub-machine to execute action sequences for certain class of external environment states. Preliminary experiments demonstrated the prospects of the method.
\end{abstract}

\section{Introduction}
\label{intro}
Hierarchical reinforcement learning (HRL) is a promising approach for solving problems with a large state space and a lack of immediate reinforcement signal \cite{Dayan1993,Wiering1997}. Hierarchical approach allows to decompose the complex task into a set of sub-tasks using hierarchical structures. It is is a natural procedure also performed by humans \cite{Rasmussen2017a}. However there is one aspect of human problem-solving that remains poorly understood --- the ability of finding an appropriate hierarchical structure. Finding good decompositions is usually an art-form and it is a major challenge to be able to automatically identify the required decomposition. Despite the fact that a number of achievements have been made in this direction \cite{Hengst2012a} discovering hierarchical structure is still open problem in reinforcement learning.

Most of the efforts aimed at learning in hierarchies are concerned acceleration of the Q-learning by identifying bottlenecks in the state space \cite{Menache2002}. The most popular framework in these works is Options \cite{Precup1998}. Within it artificial agents are able to construct and extend hierarchies of reusable skills or meta-actions (options). A suitable set of skills can help improve an agent’s efficiency in learning to solve difficult problems. Another commonly used approach in HRL is  the MAXQ framework \cite{Dietterich2000} where the value function is decomposed over the task hierarchy. Automated discovery of options hierarchy \cite{Mannor2004} and task decomposition within MAXQ approach \cite{Mehta2008b} showed good results in a number of synthetic problems e.g. Rooms or Taxi environments. Most of the real problems in robotics are very different from these artificial examples. The tasks of manipulator control and robot movement in space are of great practical interest \cite{Tamar2016,Gupta2017}. Although the existing attempts to adapt these approaches to continuous space \cite{Daniel2016}, they are of little use in these tasks. There are at least two reasons for this. The fist is a lack of mixed action and state abstraction \cite{Konidaris2016}. The second is that pseudo-rewards should be specified to learn hierarchically optimal policies.

In this paper, we focus on the automatically discovering sub-tasks and hierarchies of meta-actions within on-model variant of the HAM framework \cite{Parr1998}. One motivation for using abstract machines is that HAM approach is able to design good controllers that will realize specific behaviours. This is especially important when developing control systems for robotic systems. Among other things HAMs are a way to partially specify procedural knowledge to transform an Markov decision process (MDP) to a reduced semi Markov Decision Process (SMDP). In this paper we propose a new approach to the problem of learning structure of an abstract machine by introducing the ``internal'' environment where a state represents the structure of HAMs. We find the structure of machines for particular class of external environment states and then combine constructed machines into a superior machine. Automated discovery of such structures is compelling for at least two reasons. First, it avoids the significant human effort in engineering the task-subtask structural decomposition. Second it enables significant transfer of learned structural knowledge from one domain to the other.

\section{Background}
\subsection{Semi Markov Decision Problems}
The set of actions in hierarchical reinforcement learning consists of the primitive actions and temporal delayed or abstract actions (skills or meta-actions). Because of this, we need to extend the notation of Markov Decision Processes (MDPs) that is defined the environment in classical reinforcement learning. MDPs that include abstract actions are called semi Markov Decision Problems, or SMDPs \cite{Puterman1994}. In the task of reinforcement learning the agent's goal is to find the optimal strategy. An agent using Q-learning in the MDP environment achieves the goal by performing updates of $Q$ value, going into state $s'$ and receiving a reward $r$, after calling the action in the state $s$:

\[
    Q(s,a) \leftarrow (1-\alpha)Q(s,a) + \alpha(r+\gamma\max_{a'}Q(s',a')),
\]
where $\alpha$ is the learning rate and $\gamma$ is the discounting factor.

Let $N$ be a number of steps which are needed to complete the abstract action $a$ starting from the state $s$ and terminating in state $s'$. The transition function $T: S\times A\times S \times N\rightarrow [0,1]$ gives the probability of the action $a$: $T(s,a,s',N)=\Pr(s_{t+N}=s'|s_t=s,a_t=a)$. Then the formula of Q-learning for the SMDP is written as follows

\[
    Q(s,a) \leftarrow (1-\alpha)Q(s,a) + \alpha(r_c+\gamma^\tau\max_{a'}Q(s',a')),
\]
where $\tau$ is a number of steps performed after calling the action $a$ in the state $s$ before the state $s'$ was reached, $r_c$ - the cumulative reward received during this time.

Introducing abstract actions is important step that allows us to accelerate learning process although we continue using primitive actions. Abstract actions and SMDPs naturally lead to hierarchical structure of the set of actions and can be policies from smaller SMDP. It should be noted that HRL cannot guarantee in general that the optimal solution of a full problem will be necessarily found. 

\subsection{Reinforcement Learning with HAMs}
The HAM approach limits the possible actions of the agent by transitions between states of the machine \cite{Parr1998}. An example of a simple abstract automaton can be: "constantly move to the right or down." Transitions to certain states of the machine cause execution of actions in the environment, and the remaining transitions are of a technical nature and define internal logic. 

An abstract machine is a set of five elements $\langle M,\Sigma, \Lambda \mu, \delta\rangle$, where $M$ is a finite set of machine states, $\Sigma$ -  an input alphabet corresponding to the space of states of the environment, $\Lambda$ - the output alphabet of the abstract machine, $\delta(m,s_i)$ - the function of transition to the next state, with the current state $m\in M$ of the machine, and the state of the environment $s_i\in S$, $\mu(m)\in Lambda$ - output function of the machine.

The machine's states are divided into several types: \textit{Start} - this state starts the operation of any machine, \textit{Action} - in this state, the action to be taken is performed when the machine goes to this state, \textit{Choice} - if from this state there are several transitions, then the choice of the next one is stochastic, \textit{Call} - a transition to this state suspends the execution of the current machine and calls the machine specified in this state,\textit{Stop} - transition to this state stops execution of the current machine.

Due to the choice of the next state for the transition is not deterministic only in the state of \textit{Choice}, then update of $Q$ value is performed for previous $C$ and current $C'$ \textit{Choice} states. Iteration occurs taking into account the current state $s$ of the environment:

\[
    Q(s,C) \leftarrow (1-\alpha)Q(s,C) + \alpha(r_c+\gamma\max_{a'}Q(s',C'))
\]

\section{Hierarchy formation}
We solve the problem in an environment where additional parameters are added to standard information about states. It should be noted that there may be several additional features. All possible combinations of features we call clusters. E.g., for a well known blocks domain, it can be information in which part of the world the agent is located. 

Initially, the agent is trained using a training set of tasks. The training set consists of a number of tasks, in which the algorithm needs to generalize received information and then apply it to solve similar or even more complex task in the same environment.

At the first stage of the algorithm, an abstract machine is built for each cluster, which will always be called if the environment reports what is currently in the cluster. The construction takes place through the search of possible abstract machine, using a number of heuristics.

Generation and pruning of abstract machines occurs as follows:
\begin{enumerate}
    \item A list of possible vertices from which an machine can consist is determined. Each machine has a \textit{Start} and \textit{Finish} vertex, and there can be \textit{Action} vertices of various types of actions and a Choice vertices.
    \item Parameters is specified: the maximum number of vertices in the machine, number of vertices of each type.
    \item All possible ordered permutations of vertices are generated.
    \item The edges  begin to be added to the machine, taking into account the limitation of the HAM structure: the \textit{Start} state must have only one outgoing edge and can not have an incoming edges, the \textit{Stop} state must have only one incoming edge and can not have outgoing edges, the \textit{Action} state can only have one outgoing edge, each of the vertices can not have self-loops, there can not be edges from the \textit{Choice} state in the \textit{Stop} state, the \textit{Choice} state must have at least two outgoing edges.
    \item For each machine it is checked that all the vertices are in the same component and that all the vertices are reachable from the \textit{Start} vertex and there are no vertices that can not reach the \textit{Stop} state. Additionally, the absence of \textit{Choice} cycles is checked.
 
\end{enumerate}

When such a list of machines is built, the next stage of pruning the machine takes place. For this stage, it is necessary to train the algorithm, using for each cluster a standard machine $M_{std}$ that consists of single \textit{Choice} state and edges to all possible states of the \textit{Action} for give environment, this machine  corresponds to the choice of any action in each  environment state.

Checking of machine occurs as follows: a cluster is chosen for which the machine is built, an machine is taken, which must be checked and the trained machine. The learning algorithm is used for the selected machine and cluster, and for the remaining clusters a trained standard machine is used. If after a small number of iterations the process converges, then we memorize that the machine is applicable for current cluster.

When for each cluster a list of applicable machine is built we apply the internal environment algorithm for each of the sets of cluster machines.
According to the structure of graphs, an internal environment is formed in which the actions of the agent are the selection of the vertices of the graph (in the first step) and the addition of edges to subsequent ones. The environment is organized so that the agent's choice of action leads to the transition to the next state for which there is a proven machine.

Iterative process is started in which for each cluster an attempt is made to build a better one than the current machine with the trained internal environment. Initially, a $M_{std}$ machine is used for each of the clusters.
If the machine constructed by the internal environment leads to a better result, then it is added to the solution, otherwise the algorithm proceeds to the next step.

\begin{figure}[ht]
    \vskip 0.2in
    \begin{center}
        \centerline{\includegraphics[width=0.6\columnwidth]{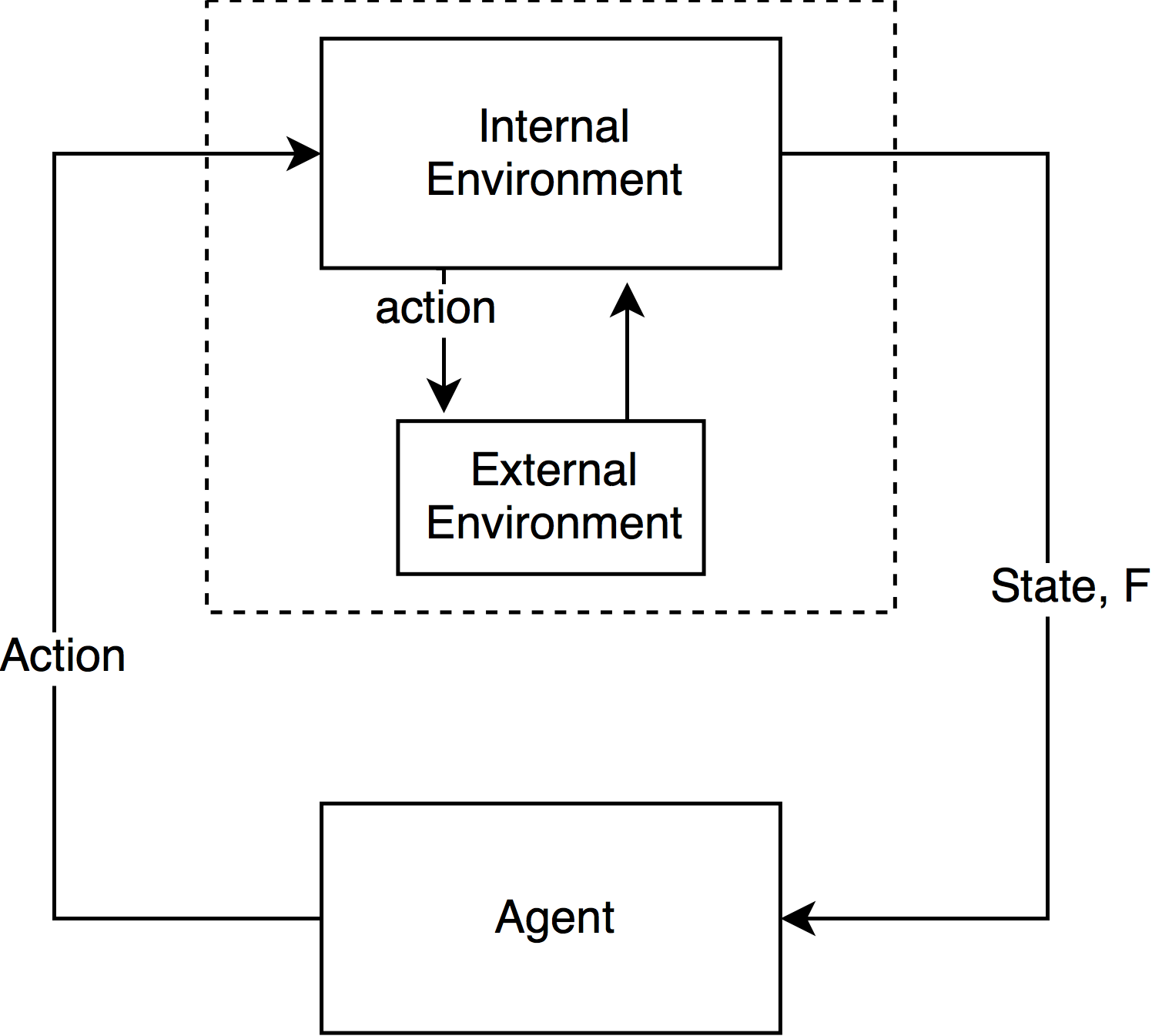}}
        \caption{Schema of the acting process  of the agent in internal or ``mental'' environment and in external or ``objective'' one.}
        \label{fig:invextenv}
    \end{center}
    \vskip -0.2in
\end{figure}

We consider the process of changing of HAM structure as a sequence of special internal or ``mental'' actions of the agent. It means that the agent acts in the second internal environment not just in external environment of surrounding objects (see fig.~\ref{fig:invextenv}). 

The state of the internal environment we assume to be a structure of a graph and some additional information containing e.g. graphs' statistics. Acting in variety of task in the external environment the agent can learn to build suitable hierarchies for the whole set of tasks in cluster. Also, the agent will try to produce hierarchies for new, previously not appearing, tasks. 

Let consider the set of tasks $\{T_1, T_2,\dots, T_k\}$ in one or different external environments in single cluster. For which task the agent will automatically built the hierarchy of HAMs. Each task $T_i$ corresponds to some SMDP for the external environment $E_{ext}$ with the set of states $S^{ext}_j$ and set of actions $A_j$.

At each agent step, the external environment returns information about the reward received at the current step $r_c$, the current state of the external environment $s_c$, information about the end of the current task $d_c$. We define an internal environment $E_{int}$ consisting of:
\begin{itemize}
    \item the set of state $S^{int}= \{s^{int}_i\}$, where each state corresponds to the structure of the graph that defines the HAM hierarchy;
    \item the set of actions $A^{int}=\{a^{int}_i\}$, where each action corresponds to some change in the structure of the graph.
\end{itemize}
The internal environment is episodic and at the beginning of each episode the agent receives information about the external environment $e_i$ for which it will be trained to build a HAM hierarchy. $N_{steps}$ is a number of steps in the internal environment. Each action performed by the agent in the internal environment causes the graph structure to change.

Consider the process of Q-learning in the internal environment:
\begin{itemize}
    \item $F_i$ a value of function that compares the statistical indicators of training with the current hierarchy to a certain number. The value of the function is calculated after training on the environment. Such an indicator can be a binary value, which is true if the agent collects the necessary total remuneration in the external environment. In this case, if the value is true, then the algorithm can decide on a possible transition to the same state of the hierarchy. Otherwise it will be advantageous to continue the search for the hierarchy;
    \item $s^{int}_{i-1}$ is a state of the HAM hierarchy in the previous step;
    \item $F_{i-1}$ -- the value of the $F$ function in the previous step;
    \item $r^{int}$  is a reward received by the agent in the previous step corresponding to the total reward $R_i$ for the external environment $E^{ext}$ within the task $T_i$. Since the learning process is sufficiently noisy, we perform several tests and take the average;
    \item $a^{int}_{i-1}$ -- the action selected in the previous step;
    \item $s_i$ -- state of the environment in the current step.
\end{itemize}
    
Then the Q-learning function will be written as usual:

\begin{multline*}
Q(s^{int}_i, a^{int}_i, F_i)\leftarrow(1- \alpha)Q(s^{int}_i, a^{int}_i, F_i) + \\\alpha\Big( r^{int} + \gamma\max_{a^{ext}_{i-1}, F_{i-1}} Q(s^{int}_{i-1}, a^{int}_{i-1}, F_{i-1})\Big)
\end{multline*}

The listing \ref{alg:intEnv} shows pseudo code of the algorithm of acting in the internal environment and illustrates the idea of transferring of total reward received in the external environment into internal reward indicating the quality of the performed ``mental'' action.

The function of transitioning to the next state of the internal environment training algorithm receives an action $ action $ acting on the input, according to which it changes the current state of the HAM, using the function $ modify $. Based on the received state of the machine, the new state is calculated, the function is $ update \_state $. It is checked whether it is possible to start a new machine using the function check for reachability of the \textit{Stop} $ check \_stop $ state and the check function for $ check \_loops $ cycles. If the test passes, then the machine starts on the outside environment and the total reward remains, otherwise the machine is associated with a large negative reward. By the state of the automaton and the reward received, the function $ F $ is calculated. 
    \SetKwInOut{Parameter}{parameter}

\begin{algorithm}
    \caption{Algorithm of the internal environment}
    \label{alg:intEnv}
    \newcommand\textbfit[1]{\bfseries\itshape{#1}}
    \SetFuncSty{textbf}
    \SetDataSty{textbfit}

    \SetKwFunction{type}{type}
    \SetKwFunction{getStart}{getStart}
    \SetKwFunction{modify}{modify} 
    \SetKwFunction{stopDistinct}{checkStop}
    \SetKwFunction{noChooseLoops}{checkLoops}
    \SetKwFunction{getPossibleActions}{getPActions}
    \SetKwFunction{getReward}{getReward}
    \SetKwFunction{runHAM}{runMachine}
    \SetKwFunction{internalStep}{internalStep}
    \SetKwFunction{updateState}{updateState}
    \SetKwFunction{F}{getF}
    \SetKwFunction{run}{run}

    \SetKwData{choose}{choose}
    \SetKwData{action}{action}
    \SetKwData{call}{call}
    \SetKwData{start}{start}
    \SetKwData{stop}{stop}
    \SetKwData{machine}{$A$}
    \SetKwData{AbstractMachine}{AbstractMachine}

    \SetKwData{nextVertex}{nextVertex}

    \SetKw{True}{True}
    \SetKw{continue}{continue}
    \SetKw{break}{break}
    \SetKw{and}{and}
    \SetKw{self}{self}
    \SetKw{return}{return}
    \SetKw{in}{in}

    \SetKwIF{If}{ElseIf}{Else}{if}{:}{elif}{else:}{}%

    \SetKwProg{Def}{def}{:}{}
    
    \Def{\internalStep(action)} {
        $\self.\machine \leftarrow \self.\machine.modify(action)$\;

        \If{\stopDistinct(\self.\machine) \and \noChooseLoops(\self.\machine)} {
            $r_{i}^{int} \leftarrow 0$\;
            \For{episode \in [0, $N_{episodes}$]}{
                $r_{i}^{int} \leftarrow r_{i}^{int} + \runHAM(\self.\machine, E^{ext})$\;
            }
        }
        \Else{
            $r_{i}^{int} \leftarrow {-\infty}$\;
        }
        $F_i$ = \F(\self.\machine, $r_{i}^{int}$)\;
        $a^{int}_{i+1} \leftarrow \getPossibleActions(\self.\machine)$\;
        
        \return $a^{int}_{i+1}$, \self.$s^{int}$, $r_{i}^{int}$, $F_i$\;
    }

    \Def{\runHAM(\machine, $E^{ext}$)} {
        $V \leftarrow \machine.getStart()$\;
        \While{$\type(V) \neq \stop$} {
            $V \leftarrow V.run()$\;
            \If{\type$(V)$ = \action} {
                $r^{ext} \leftarrow r^{ext} + V.getReward()$\;
            }
        }
        \return $r^{ext}$\;
    }

\end{algorithm}

\section{Experimental evaluation}
\begin{figure}[ht]
    \vskip 0.2in
    \begin{center}
        \centerline{\includegraphics[width=\columnwidth]{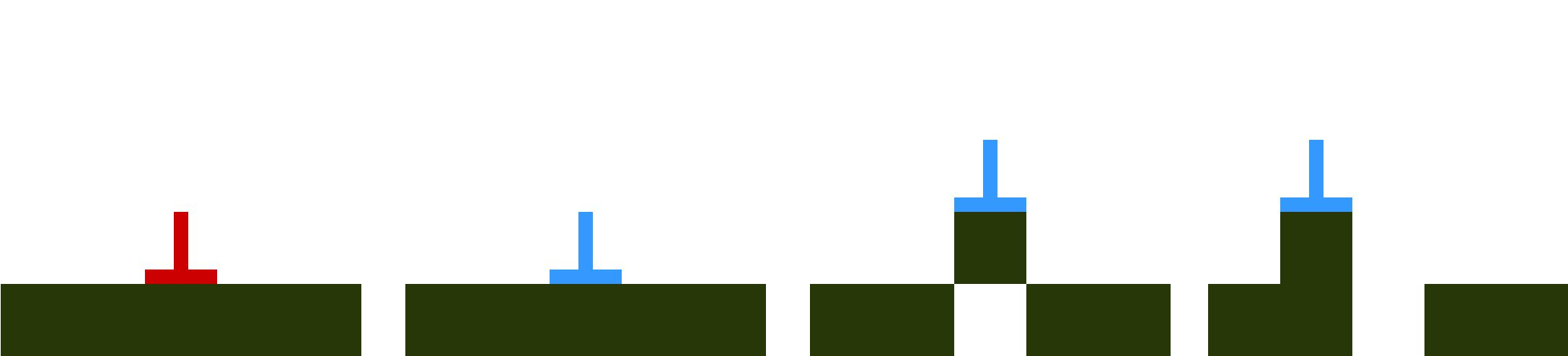}}
        \caption{An example of performing several actions in  Blocks environment (the red color of the manipulator means that the magnet is off)}
        \label{fig:blocks}
    \end{center}
    \vskip -0.2in
\end{figure}

We consider the robotic inspired environment in which a manipulator with a magnet performs actions on metal cubes. The goal is to build a tower, given the height. The agent is available 5 actions, it can move the manipulator to one unoccupied cell, in each of the four adjacent sides, and can switch the toggle of magnet (see fig.~\ref{fig:blocks}). If the magnet is turned off, then the cube instantly falls down on an unoccupied cell. Holding the manipulator, the cube moves horizontally (move left or right) only from the uppermost position. If the agent tries to apply the action, not in the upper position, then the position of the manipulator does not change, and the cube immediately drops down.

The environment is episodic and ends after a certain number of actions. The environment is completed ahead of schedule if a tower of the required height is built. For the construction of the tower of the desired height, the reward is $100$, for any other action the reward is $-0.00001$.

The division into clusters in an environment occurs according to the following parameters: at what height is the manipulator located and does the magnet holds the cube?

For the training set, we used two environments:
\begin{enumerate}
    \item height: 4, width:3, number of cubes: 3, episode length: 200, tower target size: 3.
    \item height: 5, width:4, number of cubes: 4, episode length: 500, tower target size: 4. 
\end{enumerate}

The test set consisted of one environment: height: 6, width:4, number of cubes: 5, episode length: 800, cubes tower target size: 5. 

To determine the overall reward for several environments, we used normalization with respect to the maximum reward received in the given environment with the cluster selected. For exploration, we used $\epsilon$-greedy, with initial value 0.1. The discount factor was set to 0.99. $\alpha$  was set to 0.1.

During the preliminary experiments, the above approach has not been fully demonstrated. For the stage of combining machines we used a set of the best machines built during the training. For the consolidation stage, we used a simplified algorithm. An iterative process is under way to improve the integrated solution, at each step the remaining cluster with the best total reward $R_{i}$ is taken. And two cases are checked: in the first one, the automaton of the cluster under consideration is added to the combined solution, but in the second one there is n

\begin{figure}[ht]
    \vskip 0.2in
    \begin{center}
        \centerline{\includegraphics[width=\columnwidth]{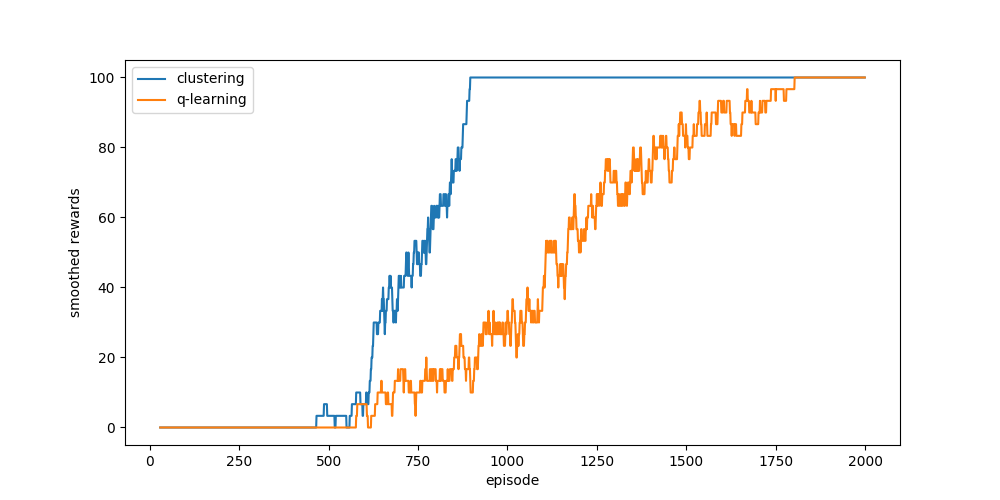}}
        \caption{Comparison of the  convergence rate of algorithms}
        \label{fig:diagram}
    \end{center}
    \vskip -0.2in
\end{figure}

The machines were not built for clusters that were not represented in each of the environments. The results of the experiment are shown on the diagram (see fig.~\ref{fig:diagram}). It shows that this approach significantly increased the rate of convergence in comparison with the standard Q-learning algorithm. The algorithm was built machines (see Appendix), which build the tower on the left and go to the right of the cubes. These meta-actions turned out to be profitable and significantly increased the learning speed of the algorithm.

\section{Conclusion}
In the paper we propose a new approach to hierarchy formation within the HAM framework. We chose the HAM abstraction because HAM approach is able to design good controllers that will realize specific behaviours. To do this, we introduced the so-called internal or ``mental'' environment in which the state marks the structure of the HAM hierarchy. The internal action in this environment leads to change the hierarchy of HAMs. We suggest the classical Q-learning in the internal environment which allows us to obtain an optimal hierarchy. We extends the HAM framework by adding on-model approach to select the appropriate sub-machine to execute action sequences for certain class of external environment states. Preliminary experiments  demonstrated the prospects of the method.

\section*{Acknowledgments}
This work was supported by the Russian Science Foundation (Project No. 16-11-00048).

\bibliography{hams_discovery}
\bibliographystyle{icml2018}

\section*{Appendix}\label{appndix}

\begin{figure}[ht]
    \vskip 0.2in
    \begin{center}
        \centerline{\includegraphics[width=\columnwidth]{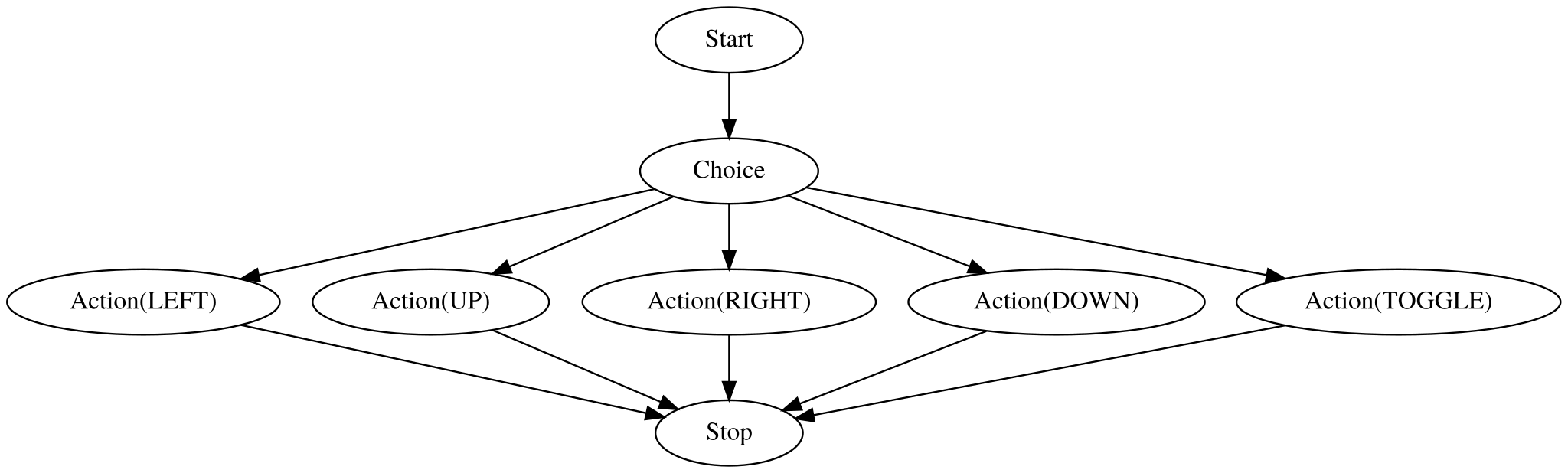}}
        \caption{An example of the constructed standard machine}
        \label{fig:stand}
    \end{center}
    \vskip -0.2in
\end{figure}

\begin{figure}[ht]
    \vskip 0.2in
    \begin{center}
        \centerline{\includegraphics[width=0.4\columnwidth]{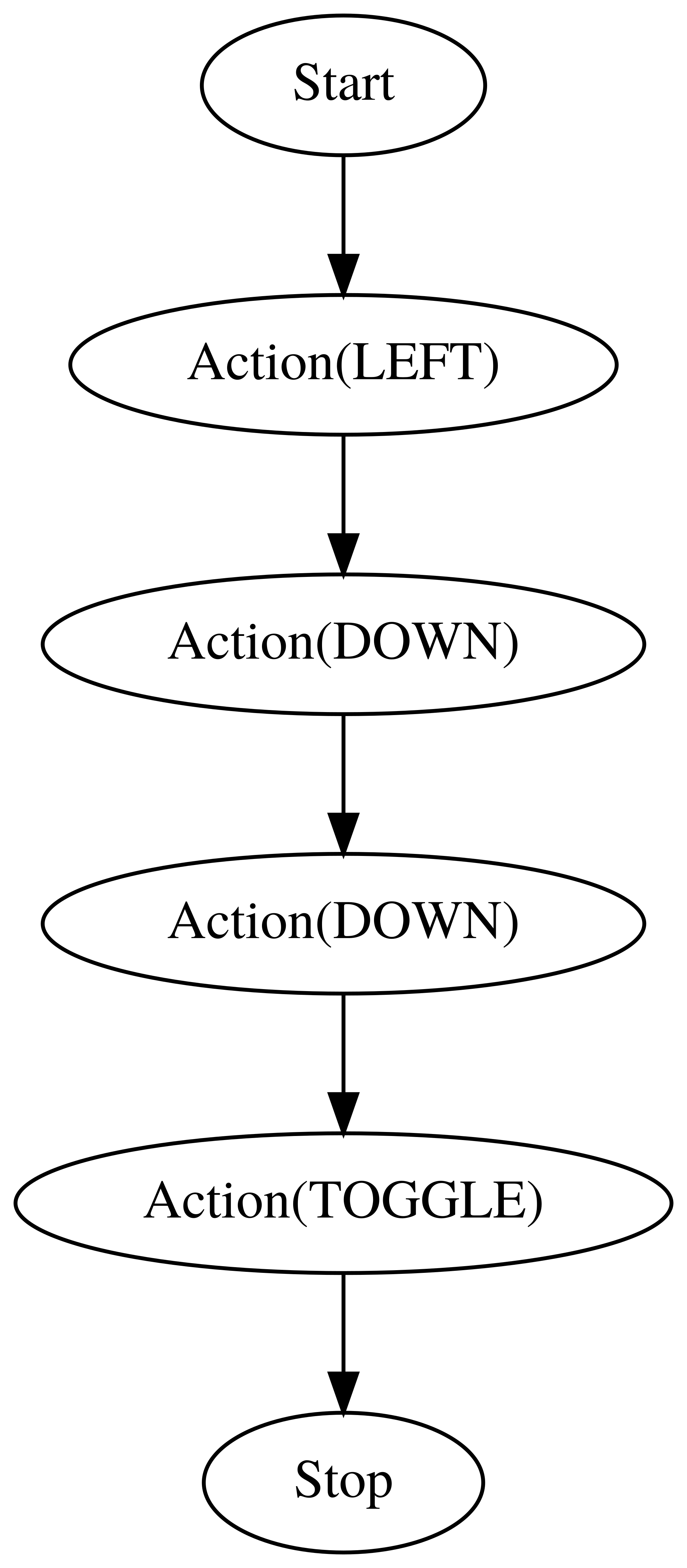}}
        \caption{An example of the constructed machine: height 0, hold is False}
        \label{fig:ex2}
    \end{center}
    \vskip -0.2in
\end{figure}

\begin{figure}[ht]
    \vskip 0.2in
    \begin{center}
        \centerline{\includegraphics[width=0.4\columnwidth]{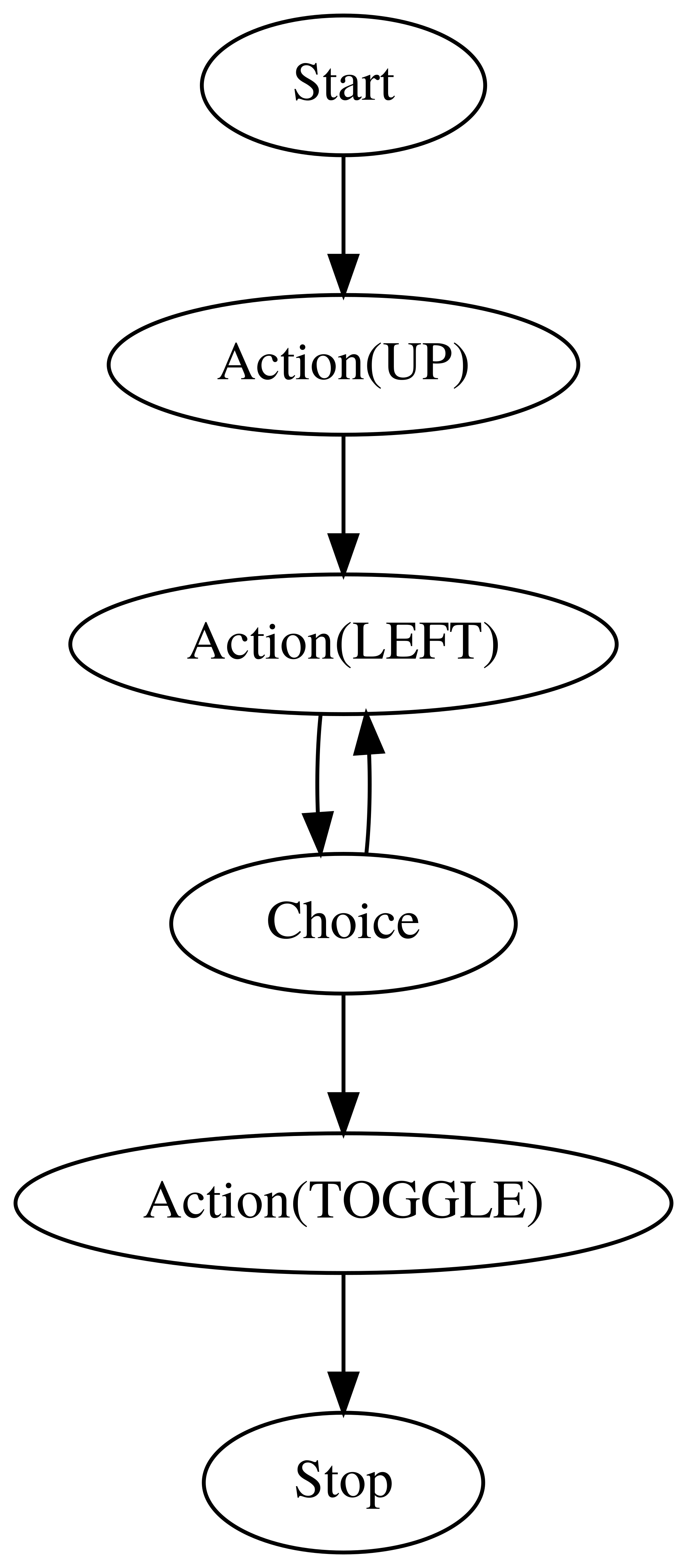}}
        \caption{An example of the constructed machine: height is 1, hold is True}
        \label{fig:ex3}
    \end{center}
    \vskip -0.2in
\end{figure}

\begin{figure}[ht]
    \vskip 0.2in
    \begin{center}
        \centerline{\includegraphics[width=0.4\columnwidth]{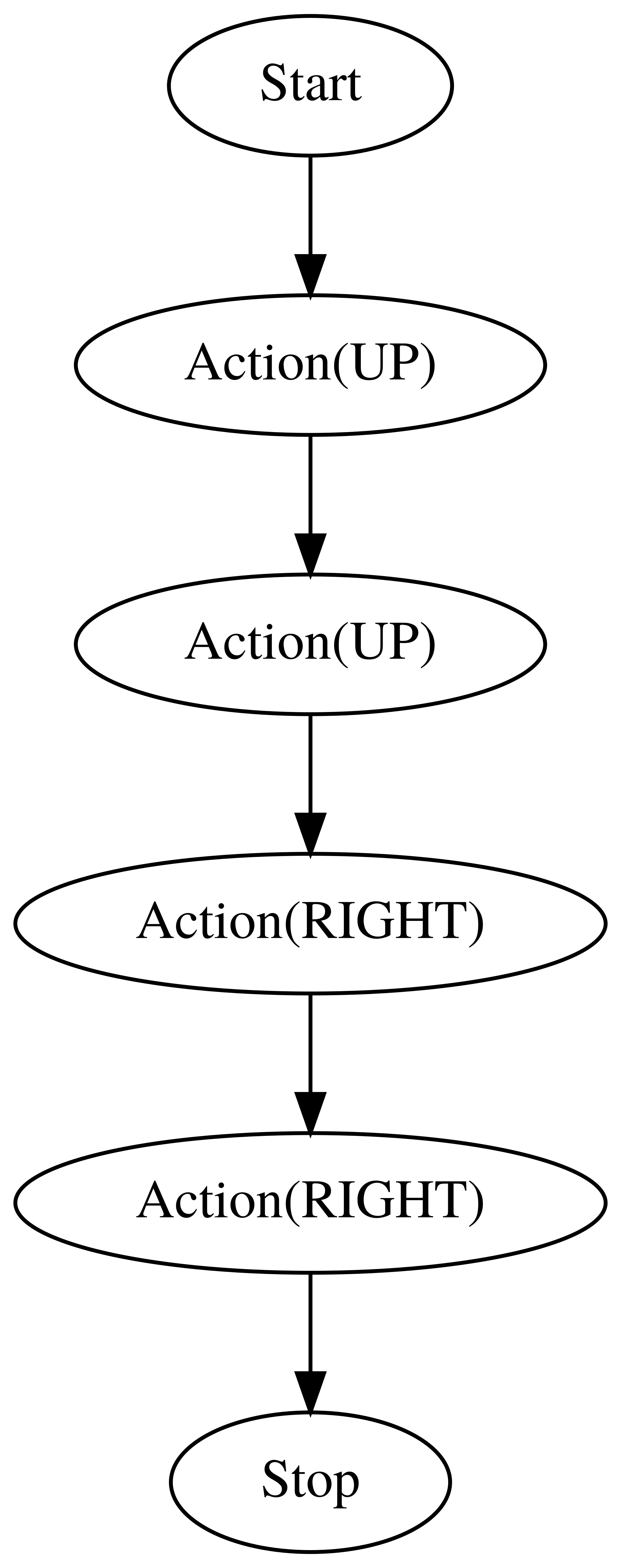}}
        \caption{An example of the constructed machine: height is 2, hold is True}
        \label{fig:ex4}
    \end{center}
    \vskip -0.2in
\end{figure}
\end{document}